\documentclass[a4paper, 10pt, conference]{ieeeconf}      
\usepackage{FG2025}

\usepackage{times}
\usepackage{epsfig}
\usepackage{amssymb}
\usepackage{latexsym}
\usepackage{float}
\usepackage{soul}
\usepackage{url}
\usepackage{amsmath}
\usepackage{algorithmic}
\usepackage[switch]{lineno}
\usepackage{pbox}
\usepackage{url}
\usepackage{color}
\usepackage{xcolor}
\usepackage{epstopdf}
\usepackage{pifont}
\usepackage{comment}
\usepackage{amsfonts}
\usepackage{tablefootnote}
\usepackage{tabularx,adjustbox}

\usepackage{times}
\usepackage{soul}
\usepackage{url}
\usepackage[hidelinks]{hyperref}
\usepackage[utf8]{inputenc}
\usepackage{graphicx}
\usepackage{booktabs}
\usepackage{algorithm}
\usepackage{algorithmic}
\usepackage[switch]{lineno}
\usepackage{color, colortbl}
\usepackage{balance}

\FGfinalcopy 

\newcommand{\name}{{\textsc{VisualCent}}}

\IEEEoverridecommandlockouts                              
\title{\LARGE \bf
{\name}: Visual Human Analysis using Dynamic Centroid Representation
}

\author{\parbox{16cm}{\centering
    {\large Niaz Ahmad$^1$ Youngmoon Lee$^2$ Guanghui Wang$^1$}\\
    {\normalsize
    $^1$ Department of Computer Science, Toronto Metropolitan University, Toronto, Canada.\\
    $^2$ Department of Robotics, Hanyang University, Ansan, South Korea.}\\
    }
}

\usepackage{fancyhdr}
\begin{document}

\maketitle

\thispagestyle{fancy}



\begin{abstract}


We introduce {\name}, a unified human pose and instance segmentation framework to address generalizability and scalability limitations to multi-person visual human analysis.
{\name} leverages centroid-based bottom-up keypoint detection paradigm
and uses \emph{Keypoint Heatmap} incorporating \emph{Disk Representation} and \emph{KeyCentroid} to identify the optimal  keypoint coordinates.  
For the unified segmentation task, an explicit keypoint is defined as a dynamic centroid called \emph{MaskCentroid} to swiftly cluster pixels to specific human instance during rapid changes in human body movement or significantly occluded environment. 
Experimental results on COCO and OCHuman datasets demonstrate {\name}'s accuracy and real-time performance advantages, outperforming existing methods in mAP scores and execution frame rate per second. The implementation is available on the project page \footnote{\url{https://sites.google.com/view/niazahmad/projects/visualcent}}.

\end{abstract}

\section{Introduction}

\label{sec:intro}

Human pose estimation and body segmentation are crucial for human-computer interaction and real-time image/video analytics. 
The primary goals are identifying individuals and their activities based on 2D joint positions and body shapes. 
The main challenges include handling an unknown number of overlapping, occluded, or entangled individuals and managing the rapid increase in computational complexity as the number of individuals grows. 
Human-to-human interactions further complicate spatial associations due to limb contacts and obstructions, necessitating an efficient, scalable, and accurate unified model for human pose and segmentation.



We propose {\name}, a new centroid-based unified representation for human pose estimation and instance-level segmentation. 
It first detects individual keypoints in a centroid-based bottom-up manner and then it employs the high confidence keypoints as dynamic centroids for mask pixels to perform instance-level segmentation.
Unlike top-down approaches \cite{ref35,ref37,ref54,han2025occluded}, {\name} detects human structures without requiring a box detector or incurring runtime complexity. 



\begin{figure}[htb!]
	\centering
	\includegraphics[width=6.5cm,height=6.6cm]{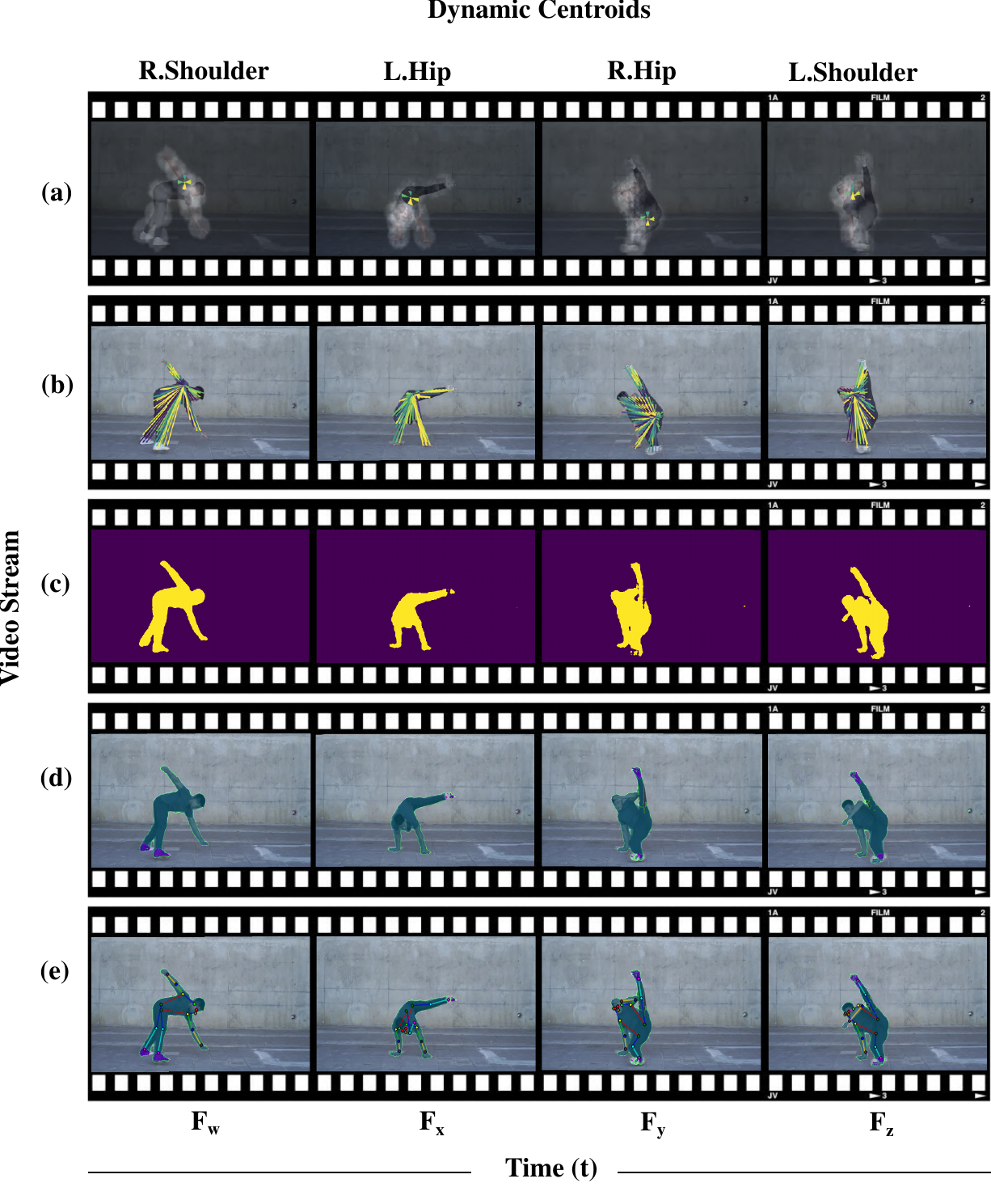}
\caption{Visual performance of {\name}. (a) represents KeyCentroid to identify precise keypoint coordinates. 
(b) represents MaskCentroid with dynamic high confident keypoint to cluster the mask pixels to the correct instance. 
(c) represents an illustration of a binary mask generated with the help of MaskCentroid 
(d) exemplifies the instance-level segmentation. 
(e) shows a unified representation of human pose and estimation (zoom in).}  
 \label{fig:video_stream}
\end{figure}

Although {\name} adopts a bottom-up approach similar to other models \cite{ref52,ref86,ref34,ref64}, it introduces significant gains. 
While existing models \cite{ref52} rely on human poses to refine pixel-wise clustering for segmentation, they struggle with segmentation accuracy. 
Other models \cite{ref86,ref34,ref64} face challenges such as i) computational overhead due to person detectors \cite{ref34}, ii) scalability limitations in instance-level segmentation \cite{ref64}, or iii) excessive model complexity \cite{ref86}, making them unsuitable for crowded scenarios and real-time applications. 
In contrast, {\name} overcomes these issues by avoiding person detectors and mitigating segmentation performance and scalability problems associated with clustering methods that lack pixel centroids.


{\name} addresses these challenges by introducing two key contributions: KeyCentroid and MaskCenroid. 
KeyCentroid is to determine the optimal keypoint coordinates with the keypoint disk by generating the vector field within the disk using a regression technique. In this way, {\name} restricts the classifier to a particular region of the keypoint that helps with intricate keypoints of the human body as shown in Fig. \ref{fig:video_stream}(a). 
MaskCenroid employs the high-confidence keypoints as dynamic clustering anchors. 
This method allows the grouping of pixels based on proximity to these centroids as shown in Fig. \ref{fig:video_stream}(b), significantly reducing the number of computations and enhancing scalability. 
MaskCentroids dynamically adjust to rapid movements and occlusions, maintaining the model's real-time capabilities without the processing overhead of exhaustive pixel-wise relationships. 
This design choice enables {\name} to operate efficiently in dense, multi-person environments without compromising accuracy or runtime performance. 
Fig. \ref{fig:video_stream}(c) illustrates the binary mask generated based on MaskCenroid concept. 
Fig. \ref{fig:video_stream}(d) shows the instance mask of the human body. 
Fig. \ref{fig:video_stream}(e) displays the final visual output of the {\name} human pose and instance-level segmentation in the video stream.

\section{Related Work}
\label{sec:related}



\noindent \textbf{Human Pose Estimation:}
Approaches for human pose estimation can be 
classified into two categories: top-down or bottom-up.
The top-down approach first runs a human detector and then 
identifies keypoints. Representative works include HRNet 
\cite{ref76}, RMPE \cite{ref36}, Multiposenet \cite{ref51}, 
Hourglass \cite{ref22}, convolutional pose machine 
\cite{ref31}, CPN \cite{ref35}, Mask r-cnn \cite{ref34}, 
simple baseline \cite{ref70},  CSM-SCARB \cite{ref80}, 
RSN \cite{ref81}, and Graph-PCNN \cite{ref82}. 
The top-down approach explores human pose within a person detector, achieving satisfactory performance but at the cost of high computational expense.
The bottom-up approach like DeepCut \cite{ref27} and 
DeeperCut \cite{ref28}, unlike the top-down counterpart, 
detects the keypoint in a one-shot manner. 
It formulates the association between keypoints as 
an integer linear scheme which takes a longer processing 
time. Part-affinity field techniques like OpenPose 
\cite{ref32} and other extensions, such as PersonLab 
\cite{ref52}, and HGG \cite{ref74} have been developed 
based on grouping techniques that often fail in the crowd. 
{\name} aims to specifically improve hard 
keypoint detection in crowded and occluded cases 
by introducing the keypoint heatmaps using keypoint 
disks and KeyCentroid.

\noindent \textbf{Instance-level Segmentation:}   
Instance-level segmentation is done in either single-stage 
\cite{ref57,ref58,ref59, wu2019unsupervised} or multi-stage \cite{ref34,ref39}. 
The single-stage approach generates intermediate and 
distributed feature maps based on the input image.
InstanceFCN \cite{ref57} produces instance-sensitive scoring 
maps and applies the assembly module to the output instance. 
This approach is based on repooling and other non-trivial 
computations (e.g., mask voting), which is not suitable 
for real-time processing. 
YOLACT \cite{ref59} generates a set of mask prototypes and 
uses coefficient masks, but this method is critical 
for obtaining a high-resolution output. 
The multi-stage approach follows the detect-then-segment 
paradigm. It first performs box detection, and then pixels 
are classified to obtain the final mask in the box region. 
Mask R-CNN \cite{ref34} 
extends Faster R-CNN \cite{ref39} by 
adding a branch for predicting segmentation masks for 
each Region of Interest. 
The subsequent work in \cite{ref68} improves the accuracy 
of Mask R-CNN by enriching the Feature Pyramid Network 
\cite{lin2017feature}. 
In contrast, our segmentation pipeline introduces 
MaskCentroid, a dynamic clustering point that helps  
cluster the mask pixels to a particular instance under 
the rapid changes in human-body movements.  

\begin{figure}[t]
	\centering
	\includegraphics[width=6.5cm,height=4.2cm]{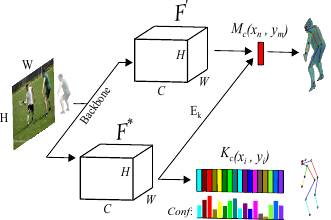}
\caption{The overview of the proposed system. \( F^* \) indicates the KeyCentroid \( K_c \) operation. \( F' \) indicates the MaskCentroid \( M_c \) operation using the Explicit keypoint $E_k$ predicted by \( K_c \).}

 \label{fig:main}
\end{figure}

\noindent \textbf{Visual Human Analysis:}  
%
The paradigm has attracted significant attention in recent years, especially in the domains of human visual analysis 
and image generation \cite{xu2024disentangled}. Mask R-CNN \cite{ref34} was a pioneering method in human visual analysis, but it suffers from high computational costs due to its top-down nature.
%
PersonLab \cite{ref52} and Pose\textit{Plus}Seg 
\cite{ref86} are closest to {\name}. 
Both of them can be considered as end-to-end joint pose and instance-level segmentation models that explore the bottom-up approach.
However, there are several major differences that make {\name} more effective, scalable, and real-time.
First, they rely on static features to detect or group keypoints by using greedy decoding; in contrast, {\name} introduces KeyCentroid that calculates the optimal keypoint coordinates, and employs MaskCentroid, a dynamic clustering point for instance-level segmentation.
Second, their segmentation does not perform well on highly entangled instances due to part-induced geometric embedding descriptors. 
Finally, they involve complex structure model with a couple of refined networks, making them infeasible for real-time applications.
\section{Technical Approach}

\noindent \textbf{Keypoint Heatmap using Disk Representation:}
\label{subsec:SKHM}
{\name} employs Keypoint Heatmap using Disk Representation (KHDR) as the foundation for human pose estimation. In this phase, individual keypoints are detected and aggregated into output feature maps. A residual-based network is adopted for multi-person pose estimation, generating keypoint heatmaps—one channel per keypoint—and KeyCentroid, two channels per keypoint for vertical and horizontal displacement within the keypoint disk.  

The keypoint prediction framework is formulated as follows: Let \( p_i \) denote a keypoint position in the image, where \( i \in \{1, \dots, N \} \) corresponds to the 2D pixel locations. A keypoint disk \( D_R(q) = \{ p : \| p - q \| \leq R \} \) with radius \( R \) is centered at \( q \), defining the local keypoint region. Similarly, let \( q_{j,k} \) represent the 2D position of the \( j \)th keypoint of the \( k \)th person instance, where \( j \in \{1, \dots, I \} \) and \( I \) is the total number of keypoints per person.  

A binary classification strategy is applied to each known keypoint \( j \), where a pixel \( p_i \) is classified as:  
\[
p_i =
\begin{cases} 
1, & \text{if } p_i \in D_R \text{ for a keypoint } j \\ 
0, & \text{otherwise} 
\end{cases}
\]

Independent dense binary classification tasks are performed for each keypoint, leading to distinct keypoint-specific heatmaps. The heatmap loss is computed using the binary cross-entropy (logistic loss) function, defined as:  

\[
\mathcal{L}_{\text{heatmap}} = - \frac{1}{N} \sum_{i=1}^{N} \left[ y_i \log(\hat{y}_i) + (1 - y_i) \log(1 - \hat{y}_i) \right],
\]
where \( N \) is the total number of pixels, \( y_i \) represents the ground-truth binary label for pixel \( p_i \), and \( \hat{y}_i \) is the predicted probability of \( p_i \) belonging to a keypoint region. This loss function quantifies the difference between predicted probabilities and true labels, and the mean loss over all pixels is used to optimize the model.

\begin{figure}[t]
	\centering
	\includegraphics[width=6.5cm,height=2.5cm]{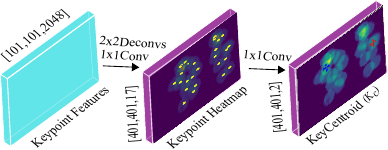}
\caption{The process of obtaining KeyCentroid involves analyzing the keypoint heatmap generated using features extracted from the backbone network.} 
 \label{fig:keycentroid}
\end{figure}



\noindent \textbf{Dynamic KeyCentroid:}  
\label{subsec:K_off}  
In addition to keypoint heatmaps, our pose operation, in conjunction with ResNet \cite{ref44}, introduces the dynamic KeyCentroid $k_{c}$ for each keypoint. The dynamic nature of $k_{c}$ allows it to adjust to variations in the spatial embedding space, ensuring robust keypoint localization even in challenging scenarios.  

The objective of the dynamic KeyCentroid is to enhance both the localization accuracy and the confidence score of keypoints. For each keypoint pixel $p_i$ within the keypoint disk $D_R$, a 2D vector $k_{v} = q_{j,k}-p_i$ originates from the image position $x$ to the $j^{th}$ keypoint of the $k^{th}$ person instance as illustrated in Fig. \ref{fig:keycentroid}. This dynamic adjustment is achieved by optimizing the centroid's position based on the spatial distribution of keypoint pixels.  

We generate multiple vector fields within $D_R$ by solving a 2D regression problem for the $j^{th}$ keypoint with spatial coordinates $(x_j, y_j)$. The response is computed on the ground truth feature map $F^{*}_{j}$ as follows:  

\begin{equation}
F^{*}_{j}(x,y) = exp\left(-\dfrac{(x-x_j)^2+(y-y_j)^2}{D_R}\right),
\end{equation}
where $D_R$ defines the keypoint disk radius, dynamically set to $R=32$ to normalize KeyCentroid and align its dynamic range with the keypoint heatmap loss. This dynamic parameterization ensures adaptability to diverse scenarios, including varying body scales and occlusions. To achieve optimal keypoint coordinates $(x_j, y_j)$, we aggregate the keypoint heatmap and the dynamic KeyCentroid, which is particularly effective for both soft and hard keypoints. This integration allows the system to maintain high precision in dynamically adapting to challenging conditions.

\begin{figure}[htb!]
	\centering
	\includegraphics[width=6.5cm,height=2.5cm]{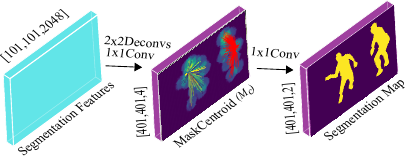}
\caption{The segmentation map is created using MaskCentroid, a dynamic keypoint that clusters mask pixels into instances with high confidence. This approach leverages segmentation features from the backbone network, ensuring precise and spatially coherent pixel assignments.} 
 \label{fig:maskcenroid}
\end{figure}

\noindent \textbf{Dynamic MaskCentroid:} 
\label{subsec:MO}
The objective of human body segmentation is to assign a set of pixels $P_i=\{m_0, m_1, m_2, ..., m_i\}$ and its 2D embedding vectors $e(m_i)$, into a set of instances $I=\{N_0, N_1, N_2, ..., N_j\}$ to generate a 2D mask for each human instance. 
Pixels are clustered to their corresponding centroid: 
%
\begin{equation}
F^{'}= \dfrac{1}{N} \sum_{m_{i} \in N_{j}} m_{i}.
\end{equation}

\noindent This is attained by defining pixel offset vector $v_i$ for each known pixel $m_i$, so that the resulting embedding $e_i=m_i+v_i$ points from its respective instance centroid. In our proposed method, a significant innovation has been made in our segmentation operation over \cite{ref91}. 
The previous model utilized the centroid as a fixed parameter to cluster the mask pixel and thus suffered inferior results if the centroid occludes in real-time cases. 
However, we can also let the network learn the optimal center of attraction by introducing the dynamic centroid. 
This can be done by defining the highly confident keypoint as a learnable parameter, as shown in Fig. \ref{fig:maskcenroid}. This approach helps in situations where rapid occlusions transpire during live video streams. The network can modify the learned parameter to adjust the center of attraction dynamically. Consequently, the network gains the ability to influence the location of the center of attraction by altering the embedding positions:

\begin{equation}
\phi_{j}(e_i)= exp\left(-\dfrac{\|e_i- {\dfrac {1}{\vert N_j \vert}} \sum_{e_{j} \in N_{j}} e_{j}\|^2}{2{\sigma}^2_j}\right).
\end{equation}

\noindent The loss functions are carefully balanced to simultaneously optimize pose estimation and instance segmentation. Specifically, the training process minimizes both the KeyCentroid loss and the MaskCentroid offset loss, while preserving the spatial relationship between keypoint positions and segmentation masks.

\section{Evaluation}
\label{sec:evaluation}

The model is evaluated on COCO \cite{ref2} and OCHuman \cite{ref64} benchmarks and trained end-to-end using the COCO keypoint and segmentation dataset. Ablation studies are conducted on the COCO validation set. ResNet-101 and ResNet-152 serve as the backbones for both training and testing. The training configuration includes a learning rate of \(0.1 \times e^{-4}\), image size \(401 \times 401\), batch size of 4, 400 epochs, and the Adam optimizer. Various transformations, including scale, flip, and rotation, are applied during training.


\begin{table}[htb!]
\centering 
   \setlength{\tabcolsep}{2pt}
    \renewcommand{\arraystretch}{0.7}
 	\fontsize{8.5}{8.5}\selectfont	
	\begin{tabular}{ l|c| c c c c c}
	\hline
    \textbf{Models} & \textbf{Backbone}& \textbf{AP} & \textbf{AP}$^{.50}$ & \textbf{AP} $^{.75}$ & \textbf{AP}${^M}$ & \textbf{AP}${^L}$ \\
	\hline
	
 
 
	
	
	
	
	OpenPose$\ast$\cite{ref105} & - & 61.8 & 84.9 & 67.5 & 57.1 & 68.2 \\
	Directpose$\ddagger$\cite{ref95} &RN-101 &64.8& 87.8& 71.1& 60.4& 71.5 \\
	A.Emb.$\ddagger$$\ast$\cite{ref3} & HG & 65.5& 86.8& 72.3& 60.6 & 72.6 \\
	PifPaf \cite{ref73} & RN-152 & 66.7 & - & - & 62.4 & 72.9 \\
	SPM \cite{ref96} &  HG & 66.9 & 88.5 & 72.9 & 62.6 & 73.1 \\

    PoseTrans\cite{ref100} &HrHR48& 67.4& 88.3& 73.9& 62.1& 75.1 \\
    
	Per.Lab$\ddagger$\cite{ref52} &RN-152 &68.7& 89.0& 75.4& 64.1& 75.5 \\
	MPose\cite{ref51} & RN-101 & 69.6& 86.3& 76.6& 65.0& 76.3 \\
    

    HrHRNet$\ddagger$\cite{ref76} & HRNet  & 70.5& 89.3& 77.2& 66.6& 75.8 \\

    PETR\cite{ref103} &SWin-L& 70.5& 91.5& 78.7& 65.2& 78.0 \\

    LOGP-CAP\cite{ref101} &HR48& 70.8& 89.7& 77.8& 66.7& 77.0 \\

     CIR\&QEM\cite{ref102} &HR48& 71.0& 90.2& 78.2& 66.2& 77.8 \\

    SIMPLE$\ddagger$ \cite{ref79} & HR32 & 71.1& 90.2& 79.4& 69.1& 79.1 \\

    Qu \textit{et al} \cite{ref99} &HrHR48& 71.1& 90.4& 78.2& 66.9& 77.2 \\

    DecentNet\cite{ref104} & HR48 & 71.2& 89.0& 78.1& 66.7& 77.8 \\

    Pose+Seg\cite{ref91} &RN-152& 72.8& 88.4& 78.7 & 67.8 & 79.4 \\

	\hline
	\hline
	
	\textbf{{\name}} &\textbf{RN-101} & \colorbox{lightgray}{74.2}& \colorbox{lightgray}{89.0}& \colorbox{lightgray}{80.2}& \colorbox{lightgray}{69.3}& \colorbox{lightgray}{81.1} \\

	\textbf{{\name}} &\textbf{RN-152} &\colorbox{lightgray}{76.1}& \colorbox{lightgray}{92.9}& \colorbox{lightgray}{83.9}& \colorbox{lightgray}{71.1}& \colorbox{lightgray}{83.5} \\
    \hline

	\end{tabular}
	
\caption{ Comparisons with \textbf{bottom-up} methods on the \textbf{COCO} test-dev 2017 set. $\ast$ denotes refinement, $\ddagger$ is multi-scale results. HG indicates Hourglass. HR indicates High-Resolution Net.}  
	
	\label{table:1}

\end{table}


\begin{table}[htb!]
\centering 
     \setlength{\tabcolsep}{9pt}
    \renewcommand{\arraystretch}{0.7}
    \fontsize{8.5}{8.5}\selectfont	
	\centering
	\begin{tabular}{ l|c| c c }
	\hline
    \textbf{Models}& \textbf{Backbone} & \textbf{Val mAP} & \textbf{Test mAP} \\
	\hline
	
	HGG $\dagger$ \cite{ref74} & HG &35.6& 34.8  \\
	HGG $\ddagger$ \cite{ref74}& HG &41.8& 36.0  \\

	MIPNet \cite{ref97} &RN-101  & 42.0& 42.5 \\

	\hline
	\hline
	\textbf{\name}& \textbf{RN-101}  & \colorbox{lightgray}{44.1}& \colorbox{lightgray}{44.6} \\
	
	\textbf{\name}& \textbf{RN-152}  & \colorbox{lightgray}{46.3}& \colorbox{lightgray}{46.0} \\
	\hline
	\end{tabular}

    \caption{Performance using \textbf{OCHuman} keypoint \textit{val} and \textit{test} datasets. $\dagger$ is single-scale and $\ddagger$ is multi-scale testing.}
	
	\label{table:key_OCHuman}
\end{table}

\noindent \textbf{Keypoint Results:}
Table \ref{table:1} summarizes the performance of {\name} on the COCO keypoint \textit{test}-dev 2017 dataset, demonstrating superior results compared to recent bottom-up methods. {\name} achieves a notable improvement of 5\% over Qu \textit{et al.} \cite{ref99}, 4.9\% over DecentNet \cite{ref104}, and 3.3\% over Pose+Seg \cite{ref91}, respectively.

Table \ref{table:key_OCHuman} shows the results of {\name} compared with state-of-the-art models on OCHuman challenging dataset. 
We assess keypoint accuracy with top competitors HGG \cite{ref74} and MIPNet \cite{ref78} both on \textit{val} and \textit{test} sets.
{\name} improves 10.0\% compare to HGG \cite{ref74} (multi-scale) and 3.5\% compare to MIPNet \cite{ref78} using \textit{test} set.

\noindent \textbf{Segmentation Results:}
Table \ref{table:4} presents segmentation results using COCO Segmentation \textit{test} sets. {\name} delivered a top accuracy of 47.6 mAP and improved the AP by 10.5\% over Mask-RCNN \cite{ref34}, 5.9\% over Per.Lab \cite{ref52} (multi-scale), and 3.1\% over Pose\textit{+}Seg \cite{ref86}. Table \ref{table:Seg_OCHuman} shows segmentation performance compared with Pose2Seg \cite{ref64} using the OCHuman \textit{val} and \textit{test} sets. 

\begin{figure}[hbt!]
    \begin{minipage}{0.48\columnwidth}
    \centering
	\includegraphics[width=4.3cm, height=3.5cm]{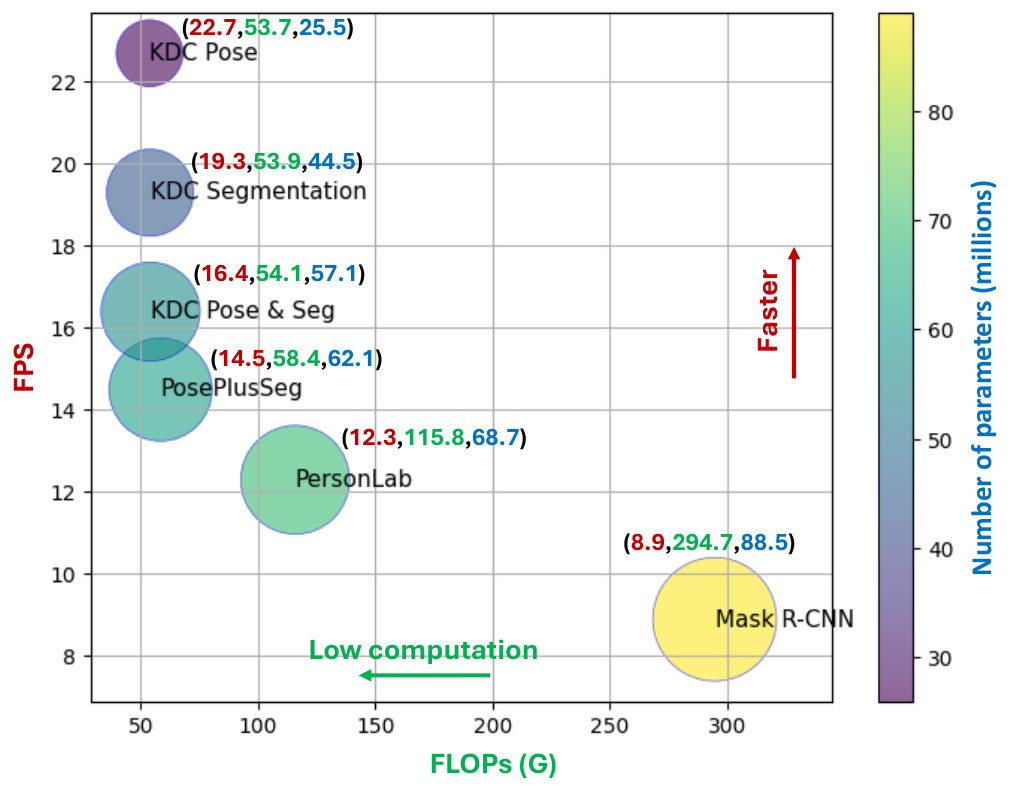}
  \caption{Computational cost with the representative sister models. Models are tested on a single Titan RTX.}
 	\label{fig:12}
	 \end{minipage}
	 \hfill
	 \begin{minipage}{0.49\columnwidth}
	\centering
		\includegraphics[width=4.6cm, height=3.6cm]{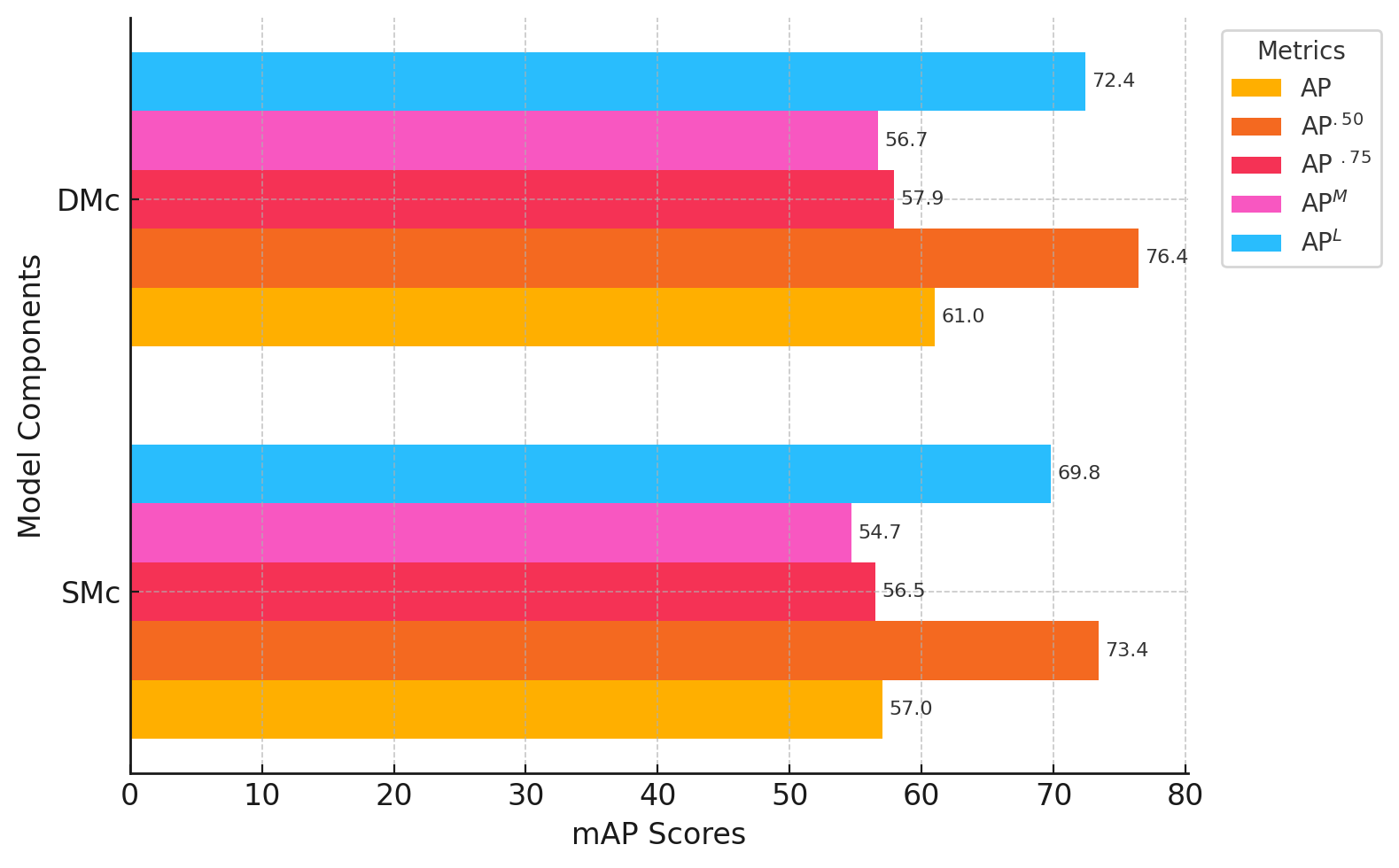}
  \caption{Performance of $SM_c$ and $DM_c$ on human instance-level segmentation.}
 	\label{fig:SMc_DMc}
    \end{minipage}
	 
\end{figure}

\noindent \textbf{Computational Cost:}
The computational cost and FPS using an image size of $401\times401$ resolution. Fig. \ref{fig:12} shows that {\name} has fewer parameters, high FPS, and lower computational complexity compared to the sister models Mask R-CNN \cite{ref34}, PersonLab \cite{ref52}, and Pose\textit{Plus}Seg \cite{ref86}.

\begin{table}[htb!]
\centering 
    \setlength{\tabcolsep}{2pt}
    \renewcommand{\arraystretch}{0.7}
    \fontsize{8.5}{8.5}\selectfont	
	\begin{tabular}{ l|c| c c c c c } 
	\hline
    \textbf{Models}& \textbf{Backbone} & \textbf{AP} & \textbf{AP}$^{.50}$ & \textbf{AP} $^{.75}$ & \textbf{AP}${^M}$ & \textbf{AP}${^L}$ \\
	\hline
	
	Mask-RCNN \cite{ref34} & RN-101 & 37.1& 60.0& 39.4& 39.9& 53.5 \\
	Per.Lab$\ddagger$\cite{ref52} & RN-101& 37.7& 65.9& 39.4& 48.0& 59.5 \\
	Per.Lab$\ddagger$\cite{ref52} &RN-152 & 38.5& 66.8& 40.4& 48.8& 60.2 \\
	Per.Lab$\ddagger$\cite{ref52} &RN-101 &41.1& 68.6& 44.5& 49.6& 62.6 \\
	Per.Lab$\ddagger$\cite{ref52} &RN-152 & 41.7& 69.1& 45.3& 50.2& 63.0 \\
	Pose+Seg\cite{ref91} & RN-152  & 44.5& 79.4& 47.1& 52.4& 65.1 \\

	\hline
	\hline
	\textbf{{\name}}& \textbf{RN-101}  & \colorbox{lightgray}{45.7}& \colorbox{lightgray}{80.4}& \colorbox{lightgray}{47.8}& \colorbox{lightgray}{53.5}& \colorbox{lightgray}{67.4} \\
	
	\textbf{{\name}}& \textbf{RN-152}  & \colorbox{lightgray}{47.6}& \colorbox{lightgray}{81.8}& \colorbox{lightgray}{48.7}& \colorbox{lightgray}{54.6}& \colorbox{lightgray}{67.8} \\
	\hline
	\end{tabular}
	\caption{Performance comparison on \textbf{COCO} Segmentation 
	\textit{test} set. $\dagger$ is single-scale  testing. $\ddagger$ is multi-scale testing.}
	\label{table:4}
\end{table}

\begin{table}[htb!]
    \setlength{\tabcolsep}{9pt}
    \renewcommand{\arraystretch}{0.7}
    \fontsize{8.5}{8.5}\selectfont	
	\label{}
	\centering
	\begin{tabular}{ l|c| c c } 
	\hline
    \textbf{Models}& \textbf{Backbone} & \textbf{Val mAP} & \textbf{Test mAP} \\
	\hline
	
	Pose2Seg \cite{ref64} &RN-50-fpn  & 54.4& 55.2 \\

	\hline
	\hline
	\textbf{\name}& \textbf{RN-101}  & \colorbox{lightgray}{56.7}& \colorbox{lightgray}{57.0} \\
	
	\textbf{\name}& \textbf{RN-152}  & \colorbox{lightgray}{58.3}& \colorbox{lightgray}{59.6}\\
	\hline
	\end{tabular}

    \caption{Comparison on \textbf{OCHuman} segmentation \textit{val} and \textit{test} datasets.}
	\label{table:Seg_OCHuman}
\end{table}

\begin{table}[htb!]

\centering 
    \setlength{\tabcolsep}{6pt}
    \renewcommand{\arraystretch}{0.7}
    \fontsize{8.5}{8.5}\selectfont	
	\begin{tabular}{ l|c c c c c}
	\hline
    \textbf{{\name}} & \textbf{AP} & \textbf{AP}$^{.50}$ & \textbf{AP} $^{.75}$ & \textbf{AP}${^M}$ & \textbf{AP}${^L}$ \\
	\hline

	\hline
        \textbf{KHDR \kern 3 em $k_{c}$}\\
	
	\textbf{\kern 1 em \checkmark} & 74.8& 89.7& 75.6 & 70.3 & 79.1\\
	\textbf{ \kern 5.9 em \checkmark}  &76.2& 91.8 & 78.9& 72.5& 82.7\\
	\textbf{\kern 1 em \checkmark \kern 4.3 em \checkmark}  &77.5& 94.9& 86.4& 73.8& 84.6\\

	\hline
	\end{tabular}

	\caption{Performance of KHDR with and without KeyCentroid $k_{c}$.}
	\label{table:5}
\end{table}



	
	





\subsection{Ablation Experiments}
\label{sec:ablation}

\noindent \textbf{KHDR and KeyCentroid:}
Initially, we evaluate the performance of the proposed KHDR and examine its effectiveness with and without the integration of $K_c$, as presented in Table \ref{table:5}. Through our ablation study, we observed that the combination of KHDR and $K_c$ is a highly effective approach for human pose estimation, particularly in challenging scenarios, enabling improved visual analysis of the human body.



\noindent \textbf{Static vs.\ Dynamic MaskCentroids:}
\label{subsec:Ab_Maskoffset}
We analyze the Static MaskCentroid ($SM_c$) and the Dynamic MaskCentroid ($DM_c$), with the results presented in Fig. \ref{fig:SMc_DMc}. The exceptional performance of the proposed $DM_c$ approach demonstrates its effectiveness in human body segmentation, particularly in scenarios involving dynamic human body movements. This capability significantly contributes to advancements in human visual analysis.

\section{Conclusion}
\label{sec:outro}


In this paper, we have introduced {\name}, a pragmatic solution for human pose estimation and instance-level human analysis. {\name} employs disk representation to generate keypoint heatmaps and KeyCentroid to obtain optimal 2D keypoint coordinates. 
Additionally, MaskCentroid is introduced as a dynamic high-confidence keypoint for clustering mask pixels in the embedding space.





\section*{ACKNOWLEDGMENTS}

\noindent This work was supported by the Institute of Information and Communications Technology Planning and Evaluation (IITP) grant IITP-2025-RS-2020-II201741, RS-2022-00155885, RS-2024-00423071, funded by the Korea government (MSIT). This work was also partly supported by the Natural Sciences and Engineering Research Council of Canada (NSERC) under grant No. RGPIN-2021-04244.




{\small
\bibliographystyle{ieee}
\bibliography{egbib}

\begin{thebibliography}{10}\itemsep=-1pt

\bibitem{ref91}
N.~Ahmad, J.~Khan, J.~Y. Kim, and Y.~Lee.
\newblock {Joint Human Pose Estimation and Instance Segmentation with PosePlusSeg}.
\newblock In {\em AAAI}, 2022.

\bibitem{ref59}
D.~Bolya, C.~Zhou, F.~Xiao, and Y.~J. Lee.
\newblock Yolact: Real-time instance segmentation.
\newblock In {\em ICCV}, 2019.

\bibitem{ref81}
Y.~Cai, Z.~Wang, Z.~Luo, B.~Yin, A.~Du, H.~Wang, X.~Zhang, X.~Zhou, E.~Zhou, and J.~Sun.
\newblock Learning delicate local representations for multi-person pose estimation.
\newblock In {\em ECCV}, 2020.

\bibitem{ref32}
Z.~Cao, T.~Simon, S.-E. Wei, and Y.~Sheikh.
\newblock Realtime multi-person 2d pose estimation using part affinity fields.
\newblock In {\em CVPR}, 2017.

\bibitem{ref105}
Z.~Cao, T.~Simon, S.-E. Wei, and Y.~Sheikh.
\newblock Realtime multi-person 2d pose estimation using part affinity fields.
\newblock In {\em Proceedings of the IEEE conference on computer vision and pattern recognition}, pages 7291--7299, 2017.

\bibitem{ref35}
Y.~Chen, Z.~Wang, Y.~Peng, Z.~Zhang, G.~Yu, and J.~Sun.
\newblock Cascaded pyramid network for multi-person pose estimation.
\newblock In {\em CVPR}, 2018.

\bibitem{ref76}
B.~Cheng, B.~Xiao, J.~Wang, H.~Shi, T.~S. Huang, and L.~Zhang.
\newblock Higherhrnet: Scale-aware representation learning for bottom-up human pose estimation.
\newblock In {\em CVPR}, 2020.

\bibitem{ref57}
J.~Dai, K.~He, Y.~Li, S.~Ren, and J.~Sun.
\newblock Instance-sensitive fully convolutional networks.
\newblock In {\em ECCV}, 2016.

\bibitem{ref86}
M.~Dantone, J.~Gall, C.~Leistner, and L.~Van~Gool.
\newblock Human pose estimation using body parts dependent joint regressors.
\newblock In {\em CVPR}, 2013.

\bibitem{ref36}
H.-S. Fang, S.~Xie, Y.-W. Tai, and C.~Lu.
\newblock Rmpe: Regional multi-person pose estimation.
\newblock In {\em ICCV}, 2017.

\bibitem{ref52}
George, Zhu, Chen, J.~Gidaris, Tompson, and K.~Murphy.
\newblock Personlab: Person pose estimation and instance segmentation with a bottom-up, part-based, geometric embedding model.
\newblock In {\em ECCV}, 2018.

\bibitem{han2025occluded}
G.~Han, C.~Song, S.~Wang, H.~Wang, E.~Chen, and G.~Wang.
\newblock Occluded human pose estimation based on limb joint augmentation.
\newblock {\em Neural Computing and Applications}, 37(3):1241--1253, 2025.

\bibitem{ref44}
He, Xiangyu, Ren, and J.~Sun.
\newblock Deep residual learning for image recognition.
\newblock In {\em CVPR}, 2016.

\bibitem{ref34}
K.~He, G.~Gkioxari, P.~Doll{\'a}r, and R.~Girshick.
\newblock Mask r-cnn.
\newblock In {\em ICCV}, 2017.

\bibitem{ref37}
S.~Huang, M.~Gong, and D.~Tao.
\newblock A coarse-fine network for keypoint localization.
\newblock In {\em ICCV}, 2017.

\bibitem{ref28}
E.~Insafutdinov, L.~Pishchulin, B.~Andres, M.~Andriluka, and B.~Schiele.
\newblock Deepercut: A deeper, stronger, and faster multi-person pose estimation model.
\newblock In {\em ECCV}, 2016.

\bibitem{ref100}
W.~Jiang, S.~Jin, W.~Liu, C.~Qian, P.~Luo, and S.~Liu.
\newblock Posetrans: A simple yet effective pose transformation augmentation for human pose estimation.
\newblock In {\em European Conference on Computer Vision}, pages 643--659. Springer, 2022.

\bibitem{ref74}
S.~Jin, W.~Liu, E.~Xie, W.~Wang, C.~Qian, W.~Ouyang, and P.~Luo.
\newblock Differentiable hierarchical graph grouping for multi-person pose estimation.
\newblock In {\em ECCV}. Springer, 2020.

\bibitem{ref78}
R.~Khirodkar, V.~Chari, A.~Agrawal, and A.~Tyagi.
\newblock Multi-hypothesis pose networks: Rethinking top-down pose estimation.
\newblock {\em arXiv preprint arXiv:2101.11223}, 2021.

\bibitem{ref97}
R.~Khirodkar, V.~Chari, A.~Agrawal, and A.~Tyagi.
\newblock Multi-instance pose networks: Rethinking top-down pose estimation.
\newblock In {\em ICCV}, 2021.

\bibitem{ref51}
M.~Kocabas, S.~Karagoz, and E.~Akbas.
\newblock Multiposenet: Fast multi-person pose estimation using pose residual network.
\newblock In {\em ECCV}, 2018.

\bibitem{ref73}
S.~Kreiss, L.~Bertoni, and A.~Alahi.
\newblock Pifpaf: Composite fields for human pose estimation.
\newblock In {\em CVPR}, 2019.

\bibitem{ref54}
W.~Li, Z.~Wang, B.~Yin, Q.~Peng, Y.~Du, T.~Xiao, G.~Yu, H.~Lu, Y.~Wei, and J.~Sun.
\newblock Rethinking on multi-stage networks for human pose estimation.
\newblock {\em arXiv preprint arXiv:1901.00148}, 2019.

\bibitem{ref2}
T.~Lin, M.~Maire, S.~Belongie, James, Perona, Deva, Piotr, and Lawrence.
\newblock Microsoft coco: Common objects in context.
\newblock In {\em ECCV}, 2014.

\bibitem{lin2017feature}
T.-Y. Lin, P.~Doll{\'a}r, R.~Girshick, K.~He, B.~Hariharan, and S.~Belongie.
\newblock Feature pyramid networks for object detection.
\newblock In {\em CVPR}, 2017.

\bibitem{ref68}
S.~Liu, L.~Qi, H.~Qin, J.~Shi, and J.~Jia.
\newblock Path aggregation network for instance segmentation.
\newblock In {\em CVPR}, 2018.

\bibitem{ref58}
J.~Long, E.~Shelhamer, and T.~Darrell.
\newblock Fully convolutional networks for semantic segmentation.
\newblock In {\em CVPR}, 2015.

\bibitem{ref3}
A.~Newell, Z.~Huang, and J.~Deng.
\newblock Associative embedding: End-to-end learning for joint detection and grouping.
\newblock In {\em NeurIPS}, 2017.

\bibitem{ref22}
A.~Newell, K.~Yang, and J.~Deng.
\newblock Stacked hourglass networks for human pose estimation.
\newblock In {\em ECCV}, 2016.

\bibitem{ref96}
X.~Nie, J.~Feng, J.~Zhang, and S.~Yan.
\newblock Single-stage multi-person pose machines.
\newblock In {\em ICCV}, 2019.

\bibitem{ref27}
L.~Pishchulin, E.~Insafutdinov, S.~Tang, B.~Andres, M.~Andriluka, P.~V. Gehler, and B.~Schiele.
\newblock Deepcut: Joint subset partition and labeling for multi person pose estimation.
\newblock In {\em CVPR}, 2016.

\bibitem{ref99}
H.~Qu, Y.~Cai, L.~G. Foo, A.~Kumar, and J.~Liu.
\newblock A characteristic function-based method for bottom-up human pose estimation.
\newblock In {\em Proceedings of the IEEE/CVF Conference on Computer Vision and Pattern Recognition}, pages 13009--13018, 2023.

\bibitem{ref39}
S.~Ren, K.~He, R.~Girshick, and J.~Sun.
\newblock Faster r-cnn: Towards real-time object detection with region proposal networks.
\newblock In {\em NeurIPS}, 2015.

\bibitem{ref103}
D.~Shi, X.~Wei, L.~Li, Y.~Ren, and W.~Tan.
\newblock End-to-end multi-person pose estimation with transformers.
\newblock In {\em Proceedings of the IEEE/CVF Conference on Computer Vision and Pattern Recognition}, pages 11069--11078, 2022.

\bibitem{ref80}
K.~Su, D.~Yu, Z.~Xu, X.~Geng, and C.~Wang.
\newblock Multi-person pose estimation with enhanced channel-wise and spatial information.
\newblock In {\em CVPR}, 2019.

\bibitem{ref95}
Z.~Tian, H.~Chen, and C.~Shen.
\newblock Directpose: Direct end-to-end multi-person pose estimation.
\newblock {\em arXiv preprint arXiv:1911.07451}, 2019.

\bibitem{ref82}
J.~Wang, X.~Long, Y.~Gao, E.~Ding, and S.~Wen.
\newblock Graph-pcnn: Two stage human pose estimation with graph pose refinement.
\newblock In {\em ECCV}, 2020.

\bibitem{ref104}
T.~Wang, L.~Jin, Z.~Wang, X.~Fan, Y.~Cheng, Y.~Teng, J.~Xing, and J.~Zhao.
\newblock Decenternet: Bottom-up human pose estimation via decentralized pose representation.
\newblock In {\em Proceedings of the 31st ACM International Conference on Multimedia}, pages 1798--1808, 2023.

\bibitem{ref31}
S.-E. Wei, V.~Ramakrishna, T.~Kanade, and Y.~Sheikh.
\newblock Convolutional pose machines.
\newblock In {\em CVPR}, 2016.

\bibitem{wu2019unsupervised}
Y.~Wu, T.~Marks, A.~Cherian, S.~Chen, C.~Feng, G.~Wang, and A.~Sullivan.
\newblock Unsupervised joint 3d object model learning and 6d pose estimation for depth-based instance segmentation.
\newblock In {\em Proceedings of the IEEE/CVF International Conference on Computer Vision Workshops}, pages 0--0, 2019.

\bibitem{ref70}
B.~Xiao, H.~Wu, and Y.~Wei.
\newblock Simple baselines for human pose estimation and tracking.
\newblock In {\em ECCV}, 2018.

\bibitem{ref102}
Y.~Xiao, D.~Yu, X.~J. Wang, L.~Jin, G.~Wang, and Q.~Zhang.
\newblock Learning quality-aware representation for multi-person pose regression.
\newblock In {\em Proceedings of the AAAI Conference on Artificial Intelligence}, volume~36, pages 2822--2830, 2022.

\bibitem{xu2024disentangled}
W.~Xu, C.~Long, Y.~Nie, and G.~Wang.
\newblock Disentangled representation learning for controllable person image generation.
\newblock {\em IEEE Transactions on Multimedia}, 26:6065--6077, 2024.

\bibitem{ref101}
N.~Xue, T.~Wu, G.-S. Xia, and L.~Zhang.
\newblock Learning local-global contextual adaptation for multi-person pose estimation.
\newblock In {\em Proceedings of the IEEE/CVF Conference on Computer Vision and Pattern Recognition}, pages 13065--13074, 2022.

\bibitem{ref79}
J.~Zhang, Z.~Zhu, J.~Lu, J.~Huang, G.~Huang, and J.~Zhou.
\newblock Simple: Single-network with mimicking and point learning for bottom-up human pose estimation.
\newblock In {\em Proceedings of the AAAI Conference on Artificial Intelligence}, volume~35, pages 3342--3350, 2021.

\bibitem{ref64}
S.-H. Zhang, R.~Li, X.~Dong, P.~Rosin, Z.~Cai, X.~Han, D.~Yang, H.~Huang, and S.-M. Hu.
\newblock Pose2seg: Detection free human instance segmentation.
\newblock In {\em CVPR}, 2019.

\end{thebibliography}
}

\end{document}